\documentclass[10pt,twocolumn,letterpaper]{article}

\usepackage{cvpr}              % To produce the CAMERA-READY version
\usepackage[dvipsnames]{xcolor}
\definecolor{cvprblue}{rgb}{0.21,0.49,0.74}
\usepackage[pagebackref,breaklinks,colorlinks,citecolor=cvprblue]{hyperref}

\usepackage{multirow}
\usepackage{float}

\usepackage{color}
\usepackage{soul}

\newcommand\blfootnote[1]{%
  \begingroup
  \renewcommand\thefootnote{}\footnote{#1}%
  \addtocounter{footnote}{-1}%
  \endgroup
}

\def\confName{CVPR}
\def\confYear{2024}

\title{HRVDA: High-Resolution Visual Document Assistant}

\author{
Chaohu Liu\textsuperscript{1,2$*\S$},
Kun Yin\textsuperscript{3$*$}, 
Haoyu Cao\textsuperscript{3}, 
Xinghua Jiang\textsuperscript{3},
Xin Li\textsuperscript{3} \\
Yinsong Liu\textsuperscript{3},
Deqiang Jiang\textsuperscript{3},
Xing Sun\textsuperscript{3},
Linli Xu\textsuperscript{1,2$\dagger$} \\
\textsuperscript{1}School of Computer Science and Technology, 
University of Science and Technology of China\\
\textsuperscript{2}State Key Laboratory of Cognitive Intelligence,
\textsuperscript{3}Tencent YouTu Lab\\
{\tt\small{liuchaohu@mail.ustc.edu.cn}}\\
{\tt\small{
\{zhanyin, rechycao, clarkjiang, fujikoli, jasonysliu, dqiangjiang, winfredsun\}@tencent.com
}}\\
{\tt\small{linlixu@ustc.edu.cn}}
}

\begin{document}
\maketitle

\blfootnote{$^*$ Equal contribution.\ \ $\dagger$ Corresponding author. \ \ $\S$ Work done during an internship at Tencent YouTu Lab.}

\begin{abstract}
Leveraging vast training data, multimodal large language models (MLLMs) have demonstrated formidable general visual comprehension capabilities and achieved remarkable performance across various tasks. 
However, their performance in visual document understanding still leaves much room for improvement. 
This discrepancy is primarily attributed to the fact that visual document understanding is a fine-grained prediction task. 
In natural scenes, MLLMs typically use low-resolution images, leading to a substantial loss of visual information.
Furthermore, general-purpose MLLMs do not excel in handling document-oriented instructions.
In this paper, we propose a High-Resolution Visual Document Assistant (\textbf{HRVDA}), which bridges the gap between MLLMs and visual document understanding.
This model employs a content filtering mechanism and an instruction filtering module to separately filter out the content-agnostic visual tokens and instruction-agnostic visual tokens, thereby achieving efficient model training and inference for high-resolution images.
In addition, we construct a document-oriented visual instruction tuning dataset and apply a multi-stage training strategy to enhance the model's document modeling capabilities.
Extensive experiments demonstrate that our model achieves state-of-the-art performance across multiple document understanding datasets, while maintaining training efficiency and inference speed comparable to low-resolution models.
\end{abstract}    
\section{Introduction}
\label{sec:intro}

Large Language Models (LLMs), such as ChatGPT~\cite{chatgpt}, LLaMA~\cite{llama, llama2}, have taken a significant stride towards general artificial intelligence. 
By leveraging massive amounts of data, they have developed powerful reasoning and instruction understanding capabilities.
The proliferation of LLMs has also faciliated the development of Multimodal Large Language Models (MLLMs), which can perceive and analyze information from images and other sources~\cite{gpt4, llava1.5, minigpt-v2, llava, minigpt-4,mPLUG-owl}.
Some existing works have demonstrated that MLLMs exhibit preliminary visual document understanding capabilities, as they can extract and comprehend information from complex documents containing textual and visual elements, such as tables, charts, and graphics~\cite{qwen-vl, cogvlm, mPLUG-DocOwl, ureader}.
Given their ability to capture the relationships between textual and visual information, employing MLLMs for visual document understanding tasks shows great potential.

\begin{figure}[t]
  \centering
   \setlength{\abovecaptionskip}{0.02cm}
   \includegraphics[width=\linewidth]{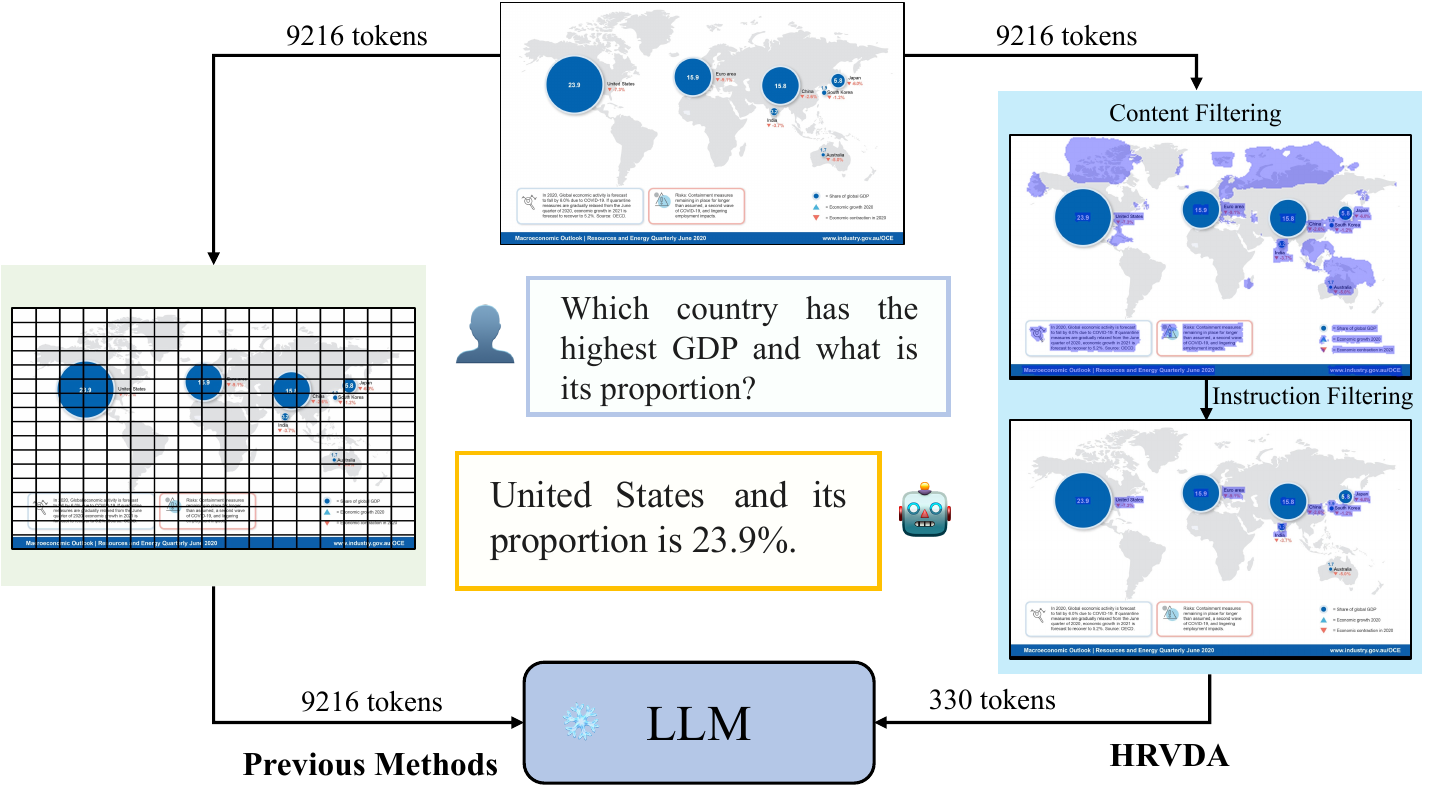}
   % \vspace{-0.5cm}
   \caption{
   Comparison of the visual processing workflow between HRVDA and previous methods.
   Previous methods generally employ a low-resolution image encoder to extract  features. 
   In contrast, HRVDA utilizes a content filtering mechanism and an instruction filtering module to selectively filter out content-agnostic and instruction-agnostic visual tokens, making high-resolution image processing computationally feasible.
   }
   \vspace{-0.5cm}
   \label{fig:motivation}
\end{figure}

However, the document image processing capabilities of MLLMs are restricted in real-world scenarios, primarily 
due to two reasons: the limitations posed by 
low-resolution image inputs and the lack of document-oriented visual instruction tuning~\cite{survey23}.

The restriction of low-resolution image inputs is a prevalent challenge in the multimodal community.
Current models usually handle images with relatively low resolutions, typically $224 \times 224$ pixels~\cite{qwen-vl, llava, minigpt-v2}. While this resolution is sufficient for
the majority of natural images, it can  result in extensive text distortion when it comes to processing document images. 
As illustrated in Figure~\ref{fig:motivation}, clear text in high-resolution images becomes blurred when resized to a lower resolution.

Directly increasing the image resolution generates a large number of visual tokens, which will occupy the limited input capacity of LLMs, and induce considerable training costs and inference latency~\cite{SparseViT}.
Taking CLIP's image encoder~\cite{clip, vit} as an example, a $1536\times 1536$ image partitioned into $16\times16$ patches results in 9216 visual tokens, which exceeds the context length of many existing open-source LLMs, such as LLaMA-2~\cite{llama2} with a context length of 4096.
In addition, they exhibit quadratic computational complexity with respect to the length of the patch sequence.

% On the other hand, general-purpose MLLMs suffer from a lack of document-oriented instruction tuning, leading to an incomplete understanding of frequently used document-oriented instructions~\cite{mPLUG-DocOwl}.
% In document analysis, elements such as layout, color, and font size often play a crucial role,  aiding in the identification and localization of specific regions.
% Consequently, there arises a pronounced demand for an augmented alignment between prevailing instructions and document images.
On the other hand, general-purpose MLLMs suffer from a lack of document-oriented visual instruction tuning~\cite{llava}, leading to an incomplete understanding of document images. 
Unlike ordinary images, document images possess distinct layout and structural information, where the font, style, and color hold significant importance for comprehending the content~\cite{docvqa, textvqa}.

% To tackle these challenges, we propose a novel multimodal large language model called \textbf{HRVDA} (\textbf{H}igh-\textbf{R}esolution \textbf{V}isual \textbf{D}ocument \textbf{A}ssistant), which employs a \textbf{content filtering mechanism} and an \textbf{instruction filtering module} designed to filter \red{out} content-agnostic visual tokens and instruction-agnostic visual tokens, respectively.

To tackle these challenges, we propose a novel multimodal large language model called \textbf{HRVDA} (\textbf{H}igh-\textbf{R}esolution \textbf{V}isual \textbf{D}ocument \textbf{A}ssistant), which employs a \textbf{content filtering mechanism} and an \textbf{instruction filtering module} designed to filter out content-agnostic visual tokens and instruction-agnostic visual tokens, respectively.

% \red{Specifically,} content-agnostic visual tokens %create
% \red{contribute} a significant amount of redundant information\red{, while the} %.
% %The
% regions in document images that contain text, tables, charts, and other document content frequently provide the most valuable information.
% As shown in Figure~\ref{fig:motivation}, the pixels within these regions constitute only a %minuscule 
% \red{small} proportion of the entire image~\cite{docvqa}.
% To reduce the number of blank background tokens, our proposed content filtering mechanism, based on a content detector, can extract key features from document images.
% Conservatively estimated, this approach filters out approximately 30\% of \blue{invalid} tokens in practice, %substantially reducing
% \red{leading to substantial reductions in } training time and inference latency without compromising performance.

Specifically, content-agnostic visual tokens contribute a significant amount of redundant information, while the regions in document images that contain text, tables, charts, and other document content frequently provide the most valuable information.
As shown in Figure~\ref{fig:motivation}, the pixels within these regions constitute only a small proportion of the entire image~\cite{docvqa}.
To reduce the number of blank background tokens, our proposed content filtering mechanism, based on a content detector, can extract key features from document images.
Conservatively estimated, this approach filters out approximately 50\% of content-agnostic tokens in practice, 
% leading to substantial reductions in training time and inference latency without compromising performance.
resulting in a substantial reduction of 30\% in training and inference latency without compromising performance.

Meanwhile, instruction-agnostic visual tokens refer to the parts that are not within the instruction attention region.
In conventional document understanding tasks, such as information extraction, document-oriented instructions often rely on localized areas to generate answers~\cite{cord, sroie}.
Therefore, we design the instruction filtering module to further filter instruction-agnostic visual tokens and significantly reduce the workload of the LLM. 

% To enhance the model's document-oriented instruction comprehension ability and achieve visual token pruning, we design a multi-stage training strategy.
% %Specifically, we initially 
% \red{In the initial stage, we} train a large image encoder using an architecture resembling that of Donut~\cite{donut}, which is a standard encoder-decoder architecture~\cite{transformer}.
% Following this, we freeze both the encoder and decoder to train the instruction filtering module. 
% Finally, we employ the Low-Rank Adaptation (LoRA) technique to align visual tokens with the LLM~\cite{lora}.
To improve the document understanding capabilities of HDVDA, we construct a document-oriented visual instruction tuning dataset. 
This dataset covers an extensive array of tasks within the document domain, including information extraction, text recognition, and visual question answering. 
It also incorporates a variety of scenarios, such as tables, charts, natural images, and webpage screenshots. 
Furthermore, we employ ChatGPT~\cite{chatgpt} to generate a diverse collection of instruction templates, thereby strengthening the generalization capabilities of the model.

Our experimental results on multiple document-oriented datasets demonstrate that HRVDA's OCR-free document comprehension capabilities surpass current state-of-the-art MLLMs such as mPLUG-DocOwl~\cite{mPLUG-DocOwl}, UReader~\cite{ureader}. % CogVLM~\cite{cogvlm}

% \noindent
In summary, our main contributions% of this paper 
are as follows:
\begin{itemize}
    \item We present \textbf{HRVDA} (\textbf{H}igh-\textbf{R}esolution \textbf{V}isual \textbf{D}ocument \textbf{A}ssistant), which, to the best of our knowledge, is the first large multimodal model designed to directly accept high-resolution image inputs.
    \item We propose a content filtering mechanism and an instruction filtering module to prune visual tokens, which significantly accelerate the model's training and inference, making the processing of high-resolution image inputs computationally feasible.
    \item We construct an extensive document-oriented visual instruction tuning dataset to enhance the model's document analysis capabilities.
    % and design a multi-stage training strategy to align visual information with the LLM. 
    \item Experimental results on a series of document-oriented datasets demonstrate that HRVDA achieves state-of-the-art performance.
\end{itemize}

\section{Related Work}

\begin{figure*}[htbp]
  \centering
  \setlength{\abovecaptionskip}{0.02cm}
   \includegraphics[width=\linewidth]{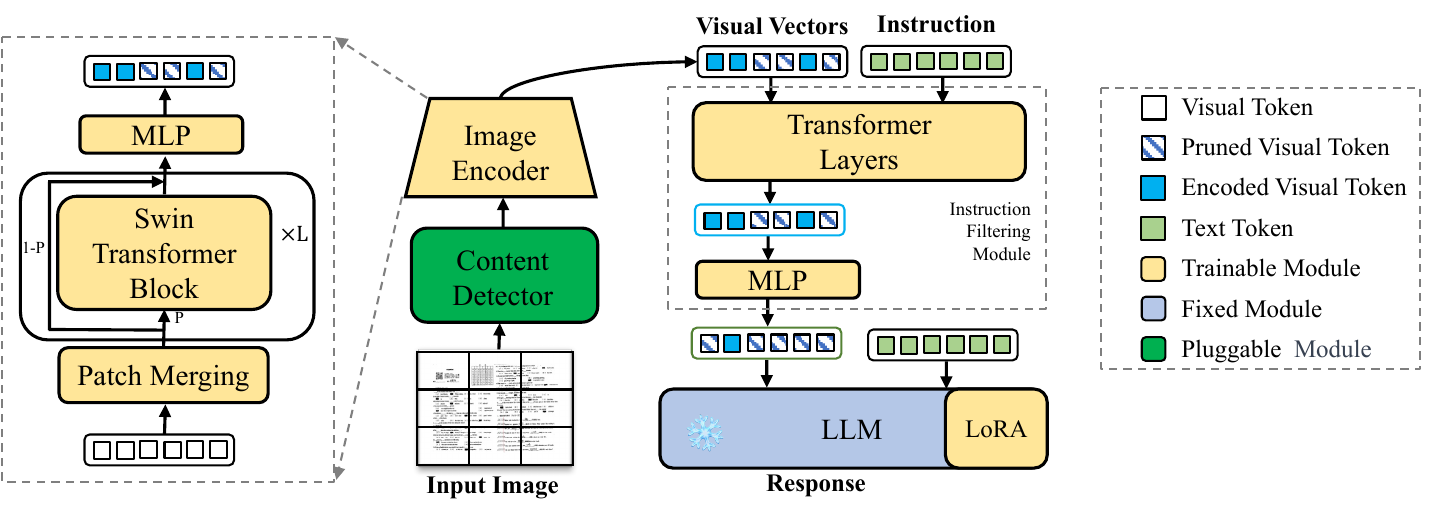}
   \caption{
%   The overall architecture of our proposed HRVDA.
%    Following a patch partition operation, a document image is transformed into an array of visual tokens. 
%    A pluggable content detector identifies whether tokens contain document content information, and then a content filtering mechanism is employed to perform token pruning. 
%    Encoded visual tokens are then processed through an MLP to maintain consistency with the LLM's embedding space dimensions.
%    The pruned token sequence is fused with the instruction features, further filtering out tokens irrelevant to the instructions. Ultimately, a highly streamlined set of visual tokens and instructions are fed into the LLM, yielding corresponding responses.
   The overall architecture of our proposed HRVDA. After partitioning the document image into visual tokens, a
    %Following a patch partition operation, a document image is transformed into an array of visual tokens. 
%    A 
pluggable content detector identifies whether tokens contain document content information, and then a content filtering mechanism is employed to perform token pruning. 
    Encoded visual tokens are then processed through an MLP to maintain consistency with the LLM's embedding space dimensions.
    The pruned token sequence is fused with the instruction features, further filtering out tokens irrelevant to the instructions. Ultimately, a streamlined set of visual tokens and instructions are fed into the LLM, generating corresponding responses.
   }
   \label{fig:model}
   \vspace{-0.2cm}
\end{figure*}

\subsection{Visual Document Understanding}
Visual Document Understanding (VDU) refers to the automated process of analyzing, comprehending, and processing document images~\cite{qgn, vdoc, wreader, xylayoutlm}.
Existing methods can be broadly categorized into two groups, OCR-dependent methods and OCR-free methods.

\textbf{OCR-dependent methods} typically rely on an external OCR interface to extract text content and coordinate information from document images~\cite{pick, chatgrid, tilt, documentai}. 
For instance, the LayoutLM family~\cite{layoutlmv1, layoutlmv2, layoutlmv3} leverages multimodal pre-training to combine image layout features with textual features.
DocFormer~\cite{docformer} undergoes unsupervised pre-training through carefully designed tasks to encourage multimodal interactions.
UDOP~\cite{udop} harmonizes image, text, and layout modalities into a unified and cohesive representation by leveraging the spatial relationships within the document.
These methods typically face issues such as increased computational costs and error accumulation~\cite{qgn}.

\textbf{OCR-free methods} aim to extract structured text directly from images in an end-to-end manner. This approach simplifies the information processing process, speeds up the reasoning and has gained significant attention in the VDU community recently~\cite{pali3, monkey}. 
For example, both Donut~\cite{donut} and Dessurt~\cite{dessurt} utilize Swin Transformer to extract image features, followed by cross-attention operations between decoder models like BART and image features to generate text in an auto-regressive manner. 
SeRum~\cite{serum} goes a step further by employing selective region concentration to enhance the precision and speed of generation.

\subsection{Multimodal Large Language Models }
MLLMs have become a new research focus recently~\cite{survey23}.
According to the modality alignment approach, they can be roughly divided into two categories: query-based methods and projection-based methods.

\textbf{Query-based methods} involve utilizing a set of learnable query tokens to extract information through cross-attention mechanisms. Flamingo~\cite{flamingo} and BLIP-2~\cite{blip2} are the first to adopt this approach, which is later inherited by a series of works~\cite{minigpt-4, instructblip, x-llm-query, video-llama-query, mPLUG-owl}. However, this method essentially introduces a textual supervisory signal to extract image features and is not suitable for fine-grained prediction tasks. The experimental results are provided in the Appendix~\ref{A.1}.

% \textbf{Projection-based methods} are widely used in the current multimodal community~\cite{pmc-vqa, llama-adapter-v2, cheap-quick, pandagpt, llava1.5}. 
\textbf{Projection-based methods} involve directly mapping visual tokens with the LLM's input space~\cite{pmc-vqa, llama-adapter-v2, cheap-quick, pandagpt, llava1.5, cogvlm}. 
For instance, LLaVA employs a simple linear layer to project image features~\cite{llava}. LLaMA-Adapter applies a lightweight adapter module to align visual tokens and text tokens~\cite{llama-adapter}.
This approach allows the LLM to perceive the entire image, offering a more promising perspective for effective multimodal learning.

% \textbf{Decoupled modality} refers to the use of an external visual expert model to obtain textual descriptions of an image and input these descriptions into LLMs.
% A typical work is MM-REACT~\cite{mmreact}, which uses a group of visual experts to obtain textual descriptions of images from multiple perspectives and inputs these descriptions to ChatGPT for multimodal reasoning.
% Such complex pipelines are subject to error transmission and lack an understanding of the fine-grained image.

% \textbf{Coupled modality} is the alignment of visual features into the input space of the LLMs, resulting in full perception of the image.
% LLaVA~\cite{llava}, PandaGPT~\cite{pandagpt} are aligned by direct projection, while BLIP-2~\cite{blip2}, MiniGPT-4~\cite{minigpt-4}  are aligned by query extraction.
% We experimentally discover that the first method is more suitable for document understanding tasks, as the latter tends to lose fine-grained information.

\subsection{Token Pruning}
Token pruning is a technique aimed at reducing model parameters and computational complexity~\cite{tokenmerging, token_Sucheng, token_tsv, SViT}. It achieves model simplification and compression by removing certain weights or feature representations.
Numerous methods for pruning vision transformers have been proposed~\cite{learning_token_pruning}. DynamicViT~\cite{DynamicViT} accelerates model inference by sparsifying less important tokens using lightweight prediction modules. SparseViT~\cite{SparseViT} efficiently processes high-resolution images through sparse activations, enabling efficient dense prediction tasks. STVit~\cite{STViT} achieves efficient global and local processing in ViTs by removing redundant image tokens and can serve as a backbone for downstream tasks.
These pruning techniques are designed for natural images and are not suitable for document images.
% This paper aims at extending token pruning methods to vision language models.
\section{HRVDA}

%In this section, we first introduce the model architecture (in Section~\ref{sec:3.1}), then provide a detailed explanation of the \textbf{Content Filtering Mechanism} (in Section~\ref{sec:3.2}) and the \textbf{Instruction Filtering Module} (in Section~\ref{sec:3.3}). Finally, we discuss the instruction tuning dataset constructed for document understanding (in Section~\ref{sec:3.4}) and the training strategy (in Section~\ref{sec:3.5}).

% In this section, we %first introduce
% \red{start with} the model architecture (in Section~\ref{sec:3.1}), %then provide
% \red{followed with} a detailed explanation of the \textbf{Content Filtering Mechanism} (in Section~\ref{sec:3.2}) and the \textbf{Instruction Filtering Module} (in Section~\ref{sec:3.3}). Finally, we %discuss 
% \red{introduce} the instruction tuning dataset constructed for document understanding (in Section~\ref{sec:3.4}) and the training strategy (in Section~\ref{sec:3.5}).

In this section, we start with the model architecture (in Section~\ref{sec:3.1}), followed with a detailed explanation of the \textbf{Content Filtering Mechanism} (in Section~\ref{sec:3.2}) and the \textbf{Instruction Filtering Module} (in Section~\ref{sec:3.3}). Finally, we introduce the instruction tuning dataset constructed for document understanding (in Section~\ref{sec:3.4}) and the training strategy (in Section~\ref{sec:3.5}).

\subsection{Overall Architecture}\label{sec:3.1}
HRVDA is a large multimodal model designed to address the challenges posed by high-resolution requirements in visual document understanding tasks.
As shown in Figure~\ref{fig:model}, it mainly consists of four modules: a content detector, an image encoder, an instruction filtering module (IFM), and an LLM.

% \red{The initial step involves partitioning the original image} %The original image is partitioned 
% into a series of patches, which are subsequently converted into a sequence of visual tokens.
% These tokens are then processed by a content detector to %ascertain
% \red{assess} the probability of each token containing significant information. 
% %By utilizing
% \red{Leveraging} these probabilities, a content filtering mechanism enables the image encoder to selectively compute visual features and eliminate content-agnostic visual tokens.
% These encoded visual features %, in conjunction with the instruction features, 
% are subsequently integrated \red{with the instruction features} using a self-attention mechanism within %an 
% \red{the} instruction filtering module. A straightforward 2-layer MLP network is employed to classify these fused features and further exclude instruction-agnostic visual tokens.
% Ultimately, the highly refined visual tokens are concatenated with the instruction tokens and %input
% \red{fed} into the LLM for generating the anticipated response. This approach ensures a more efficient and effective representation of the image content, tailored specifically for the task at hand.

The initial step involves partitioning the original image into a series of patches, which are subsequently converted into a sequence of visual tokens.
These tokens are then processed by a content detector to assess the probability of each token containing significant information. 
Leveraging these probabilities, a content filtering mechanism enables the image encoder to selectively compute visual features and eliminate content-agnostic visual tokens.
These encoded visual features are subsequently integrated with the instruction features using a self-attention mechanism within the instruction filtering module. 
A straightforward 2-layer MLP network is employed to classify these fused features and further exclude instruction-agnostic visual tokens.
Ultimately, the highly refined visual tokens are concatenated with the instruction tokens and fed into the LLM for generating the anticipated response. 
This approach ensures a more efficient and effective representation of the image content, tailored specifically for the task at hand.

%\subsection{Content Filtering Mechanism}
% \subsection{\red{Content Filtering}}
\subsection{Content Filtering}
\label{sec:3.2}

In conventional Transformer architectures~\cite{transformer}, high-resolution images are converted into long token sequences, which poses a substantial demand on computational resources. 
Moreover, elongated sequences introduce challenges in capturing long-range dependencies.

A potential solution to these challenges lies in the unique properties of document images: 
they typically consist of extensive areas of blank background, while content-rich regions provide the majority of valuable information~\cite{docvqa}. 
To leverage the sparse content information effectively and efficiently, we propose a content filtering mechanism, primarily involving two modules: the content detector and the image encoder.

% \noindent
\textbf{Content Detector.}
A pluggable network is employed to identify whether each token contains important content. 
For document images, such content includes elements such as text, tables, and charts~\cite{docvqa}.
The choice of network can be quite diverse. It could be a simple MLP network for token classification, a detection network like DETR~\cite{detr}, or a segmentation network like U-Net~\cite{unet} applied to reshaped feature maps.
In this work, we employ a shallow PSENet~\cite{PSENet}, which is designed as a segmentation-based detector capable of localizing text instances of any shape.
The content detector adopts a high recall rate strategy, ensuring that all visual tokens containing content are preserved.

% \noindent
\textbf{Image Encoder.} 
A visual backbone network is used to extract image features.
In contrast to most MLLMs that utilize ViT~\cite{vit}, we adopt the Swin Transformer~\cite{swinT} as our image encoder, which utilizes a window-based mechanism for self-attention computation, mitigating computational burdens.
Moreover, it incorporates a token merge mechanism to prevent the direct loss of information. 
% The Swin Transformer's downsampling of feature maps also contributes to %reducing
% \red{a further reduction of} the number of visual tokens.
% % Additionally, it employs a token merge mechanism to avoid the direct loss of information.
% \red{\st{In addition, its downsampling of feature maps also reduces the number of visual tokens.}}
The Swin Transformer's downsampling of feature maps also contributes to a further reduction of the number of visual tokens.
% Additionally, it employs a token merge mechanism to avoid the direct loss of information.
% \red{\st{In addition, its downsampling of feature maps also reduces the number of visual tokens.}}

Given an image $\mathrm{x} \in \mathbb{R}^{H \times W \times C}$, the patch partition module transforms it into a set of visual tokens $\{z_i \mid z_i\in \mathbb{R}^d, 1\leq i \leq n \}$, where $n$ represents the number of image patches and $d$ is the dimension of the latent vectors of the encoder. 
The content detector performs a binary classification task on the visual tokens and can obtain the probability $\{p_i \mid p_i \in [0, 1], 1 \leq i \leq n \}$ that each patch contains valuable content.
Note that the patch partition module employed by the content detector exhibits a structure similar to that of the Swin Transformer, yet they do not share parameters.

% In each Swin Transformer block, a skip connection is introduced:
As shown on the left side of Figure~\ref{fig:model}, a skip connection is introduced in each Swin Transformer block to accelerate computation:
\begin{equation}
    \begin{split}
        h^{j+1} =  p * F^j(h^j) & + (1-p) * h^j \\
        h^0 =  & z
    \end{split}
\end{equation}
where $F^j$ represents the operation in the $j$-th Swin Transformer block, and $h^j$ is the hidden state of the visual tokens.
% and $\odot$ denotes the Hadamard product.

For a well-trained content detector, we employ a threshold $\epsilon_c$ to adjust the probability values in $P$ for tokens containing content:
\begin{equation}
    p_i= \left\{\begin{array}{rcl}
    1, \quad p_i \geqslant \epsilon_c \\
    0, \quad p_i < \epsilon_c
    \end{array}\right.
\end{equation}
Utilizing these probabilities, if none of the tokens within a window %are 
is considered to contain content, the window bypasses the attention computation and is directly passed to the next block, thereby achieving computational acceleration.

It is worth highlighting once again that content-agnostic tokens are not directly removed,  making the four merging adjacent patches spatially close. The shifted window partitioning approach~\cite{swinT} in the Swin Transformer enables interactions between different tokens, thereby preserving potentially useful layout information and enhancing modeling capabilities.

The patch merging operation in Swin Transformer consolidates adjacent $2\times2$ regions into a single new patch, and the probability of the merged patch containing content is set to the maximum value among the probabilities of the 4 original regions: %the original four probabilities:
\begin{equation}
    p_i' = \max (p_i, p_{i+1}, p_{i+2}, p_{i+3})
    \label{p}
\end{equation}

To further preserve global information, the threshold value $\epsilon$ is progressively increased %for each stage varies, and the strategy in this work is to progressively increase the threshold 
from shallow to deeper layers. 
% \blue{Appropriate redundancy contributes to enhancing the model's robustness.}
Preserving more tokens in the shallow layers can reduce the loss of visual information.

\subsection{Instruction Filtering}
\label{sec:3.3}
%Document-oriented instructions are highly purposeful and typically refer only to specific regions within the image, which suggests that further filtering of visual tokens is promising.

% Document-oriented instructions are highly %purposeful
% \red{specific,} %and 
% typically %refer
% \red{referring} only to %specific
% \red{particular} regions within the image, which 
% \red{necessitates} %suggests that 
% further filtering of visual tokens. %is promising.

Document-oriented instructions are highly specific, %and 
typically %refer 
referring only to particular regions within the image, which necessitates %suggests that 
further filtering of visual tokens. %is promising.

Several existing methods, for instance, the Q-Former module in BLIP-2~\cite{blip2, instructblip} and the Visual Abstractor in mPLUG-owl~\cite{mPLUG-owl}, employ learnable queries to extract valuable information. Nevertheless, this approach inadvertently leads to a diminished representation of visual information, %rendering it unsuitable 
making it less suitable for fine-grained prediction tasks. Moreover, the inclusion
of query vectors essentially %employs
relies on text as a supervisory signal, yet the textual descriptions of images %is
are often insufficient to provide accurate representations. %for an accurate representation.
On the other hand, we experimentally discover that for high-resolution images, approximately 500 query vectors are required to maintain performance without significant degradation. 
This indicates that this approach does not offer a computational advantage in terms of processing speed.

In this study, we utilize a more direct instruction filtering module (IFM) that avoids excessive compression of visual information, thus preserving its integrity.

\begin{table}[t]
\centering
\begin{tabular}{ll}
\toprule
\textbf{Task}                 & \textbf{Format}                                     \\
\midrule
\multirow{2}{*}{DC}  & Human: What is the category of this image? \\
                     & AI: \{cls\}                                \\
\hline
\multirow{2}{*}{IE}  & Human: what is the value of the \{key\}?   \\
                     & AI: \{value\}                              \\ \hline
\multirow{2}{*}{VQA} & Human: \{question\}                        \\
                     & AI: \{answer\}                             \\ \hline
\multirow{2}{*}{OCR} & Human: Present all the text in the image.  \\
                     & AI: \{all text\}                               \\ \hline
\multirow{2}{*}{VG}  & Human: Where is the \{obj\}?               \\
                     & AI: \{x, y, x + w, y + h\}                 \\ \hline
\multirow{2}{*}{IC}  & Human: What is the abstract of the image?  \\
                     & AI: \{caption\}                            \\ \hline
\multirow{2}{*}{TR}  & Human: What is the element in the table?   \\
                     & AI: \{element\}                           \\
\bottomrule
\end{tabular}
\setlength{\abovecaptionskip}{0cm}
% \setlength{\belowcaptionskip}{-0.2cm}
%\caption{Example demonstrations of instruction tuning templates tailored to seven tasks.}
\caption{%Example demonstrations 
% \red{Illustrative examples} of instruction tuning templates %tailored to
% \red{customized for} seven tasks.
Illustrative examples of instruction tuning templates %tailored to
customized for seven tasks.
}
\label{tab:it}
\vspace{-0.5cm}
\end{table}

Formally, the visual vectors obtained from the image encoder and the instruction vectors are concatenated and then fed into the instruction filtering module for further processing.
Then, a Transformer layer is employed to facilitate the fusion of these feature vectors:
\begin{equation}
    [V', I'] = FFN(SA([V, I]))
\end{equation}
where $SA$ stands for  the self-attention layer, $FFN$ represents the feedforward layer, and $V$, $I$ denote the visual vectors and instruction vectors, respectively.
The fused visual features $V'$ are then sent to a 2-layer MLP for binary classification to filter out visual tokens that are irrelevant to the instructions~\cite{SViT}.
% Similar to the content detector, a high recall rate design is implemented in this context, ensuring that relevant information is not inadvertently discarded during the filtering process.
Similar to the content detector, the instruction filtering module also adopts a filtering threshold $\epsilon_i$, as in Equation~\ref{p}, to increase the classification recall rate, ensuring that visual tokens related to instructions are not discarded.

Ultimately, following content-agnostic and instruction-agnostic filtering, the visual token sequences are fed into the LLM.

\begin{figure}[t]
  \centering
  \setlength{\abovecaptionskip}{0.02cm}
   \includegraphics[width=\linewidth]{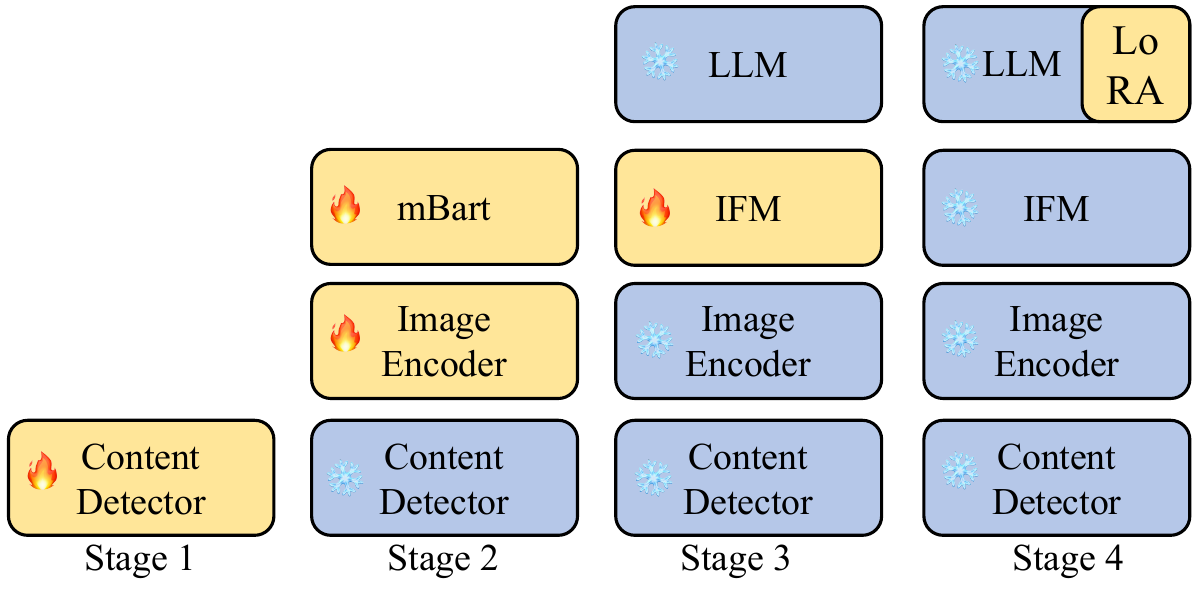}
   \caption{
   The training pipeline of our HRVDA model.
   }
   \label{fig:train}
   \vspace{-0.7cm}
\end{figure}

\subsection{Visual Instruction Tuning}
\label{sec:3.4}
In this section, we primarily introduce the task of visual instruction tuning and %their
the data sources.

\textbf{Tuning Tasks.}
To enhance HRVDA's generalization in visual document understanding, we organize a wide range of document tasks into an instruction format.
In this work, we primarily focus on tasks such as document classification (DC), information extraction (IE), visual question answering (VQA), optical character recognition (OCR), visual grounding (VG), image captioning (IC), and table reconstruction (TR).
% As shown in Table ~\ref{tab:it}, we construct some basic templates.
Table~\ref{tab:it} presents some fundamental examples.

To diversify the range of prompts, we 
first manually craft 10 prompt templates  
for each task. Subsequently, we employ ChatGPT~\cite{chatgpt} to generate 50 similar prompts, which are then reviewed by human experts to ensure their alignment with the intended meaning. Additional templates can be found in the Appendix~\ref{A.templates}.

\textbf{Instruction Data Resources.}
% A substantial amount of publicly available datasets and synthetic data are collected.
A large number of real-world and synthetic datasets are collected.
The real-world datasets used in this study include IIT-CDIP~\cite{IIT-CDIP}, CORD~\cite{cord}, SROIE~\cite{sroie}, DocVQA~\cite{docvqa}, InfographicsVQA~\cite{InfographicVQA}, DeepForm~\cite{due-deepform}, Kleister Charity~\cite{KLC}, WikiTableQuestions~\cite{WikiTableQuestions}, TabFact~\cite{TabFact}, ChartQA~\cite{ChartQA}, TextVQA~\cite{textvqa}, TextCaps~\cite{textcaps}, VisualMRC~\cite{VisualMRC}, PubTabNet~\cite{tr}, \textit{etc}.
Given the limited availability of open source data, in this work a significant amount of data synthesis methods are applied, such as SynthText~\cite{SynthText}, Synth90K~\cite{Synth90K} and SynthDoG~\cite{donut}.
Due to space constraints, more details can be seen in the Appendix~\ref{A.dataset}.

% PaLI-3 &  &  &  & 87.6 & 57.8 & & & & & 72.0 & 79.5 & & 158.8\\
\begin{table*}[htbp] 
  \centering % 居中表格
  \setlength{\tabcolsep}{1.86mm}{
  % \resizebox{\linewidth}{!}{
  \begin{tabular}{l|c|cc|cccccccccc}
    \toprule % 添加顶部线
    \multirow{2}*{Model} & \multirow{2}*{Res.}   & \multirow{2}*{CORD} & \multirow{2}*{SROIE} & Doc & Info   & Deep  & \multirow{2}*{KLC} & \multirow{2}*{WTQ}   & Tab  & Chart & Text & Visual   & Text\\
        &  & &   & VQA  & VQA & Form &   &  & Fact & QA  & VQA  & MRC & Caps \\
    \midrule
    Donut$^\ast$ & 1280 & 84.1 & 83.2 & 67.5 & 11.6 & 61.6 & 30.0 & 18.8 & 54.6 & 41.8 & 43.5 & 93.9 & 74.4\\
    SeRum$^\ast$ & 1280 & 84.9 & 85.8 & 71.9 & 13.5 & 50.7 & 31.3 & 25.5 & 58.3 & 47.9 & 66.3 & 98.6 & 101.4 \\
    Pix2Struct & 1024 & - & - & \textbf{76.6} & 40.0 & - & - & - & - & 58.6  & - & - & - \\
    \hline
    CogVLM & 490 & - & - & - & - & - & - & - & - & - & 69.7 & -  & \textbf{144.9} \\
    Qwen-VL$^\dagger$ & 448 & - & - & 65.1 & 29.9 & 2.2 & 8.9 & 16.1 & 52.5 & 66.3 & 63.8 & 76.5 &20.25\\
    mPLUG-Doc & 224 & - & - & 62.2 & 38.2 & 42.6 & 30.3 & 26.9 & 60.2 & 57.4 & 52.6 & 188.8 & 111.9 \\ % 表头，列之间用 & 分隔，\\ 表示换行
    UReader & 224 & - & - & 65.4 & 42.2 & 49.5 & 32.8 & 29.4 & 67.6 & 59.3 & 57.6 & \textbf{221.7} & 118.4 \\
    % \midrule % 添加中间线
    \textbf{HRVDA} & 1536 & \textbf{89.3} & \textbf{91.0} & 72.1 & \textbf{43.5} & \textbf{63.2} & \textbf{37.5} & \textbf{31.2} & \textbf{72.3} & \textbf{67.6} & \textbf{73.3} & 211.5 & 125.3 \\ % 第一行数据
    \bottomrule % 添加底部线
  \end{tabular}
  }
  \setlength{\abovecaptionskip}{0cm}
    \caption{
    Comparison of HRVDA with OCR-free models across 12 document domain datasets.
    For consistent comparison, $^\ast$ denotes results obtained after fine-tuning, while $^\dagger$ indicates results evaluated based on open-source models.
    The best results are marked in bold.} 
    \label{tab:main} % 为表格添加标签，以便在正文中引用
    \vspace{-0.5cm}
\end{table*}

\begin{table}
    \centering
    \begin{tabular}{lccccc}
        \toprule
        Settings     & Res.     & Encoder & Decoder & All   \\
        \midrule
        Qwen-VL      &    448   & 1.67   & 7.8   & 9.47  \\
        HRVDA(0.25, 0.25)  & 1536  & 0.92    & 6.33  & 7.25   \\
        HRVDA(0.25, 0.5)  & 1536  & 0.89    & 4.68    & 5.57  \\
        HRVDA(0.5, 0.25)  & 1536  & 0.75    & 4.05   & 4.80  \\
        HRVDA(0.5, 0.5)  & 1536   & 0.76    & 2.88   & 3.64 \\
        \bottomrule
    \end{tabular}
    \setlength{\abovecaptionskip}{0cm}
    \caption{Comparison of forward-inference efficiency between HRVDA and Qwen-VL. 
    HRVDA is configured with four sets of filtering thresholds for content and instruction.}
    \vspace{-0.5cm}
    \label{tbl:time}
\end{table}

\subsection{Training Strategies}
\label{sec:3.5}
In order to achieve visual token filtering and enhance the model's document-oriented instruction understanding capabilities, a multi-stage training strategy is adopted in this work as shown in Figure~\ref{fig:train}.

\textbf{Stage 1} focuses on training the content detector. 
We employ external OCR tools and detection networks to obtain the coordinates of various elements, including text, charts, tables, \textit{ etc.}
These coordinates can be used to provide supervised signals for the PSENet, determining whether each visual token contains content or not.
\textbf{Stage 2} concentrates on the pretraining of the image encoder.
Our encoder is integrated with m-BART~\cite{mbart} via cross-attention to perform the task of recognizing all text within the images~\cite{donut}.
\textbf{Stage 3} involves the training of the instruction filtering module.
For data with fixed layouts, a high filtering threshold is used. Conversely, we utilize a low filtering threshold for data characterized by variable layouts.
\textbf{Stage 4} entails implementing low-rank adaptation techniques to preserve the general conversational capabilities of the LLM~\cite{lora}.
Additional training details can be found in the Appendix~\ref{A.train}.
\begin{figure*}[ht]
  \centering
  \setlength{\abovecaptionskip}{0.02cm}
   \includegraphics[width=\linewidth]{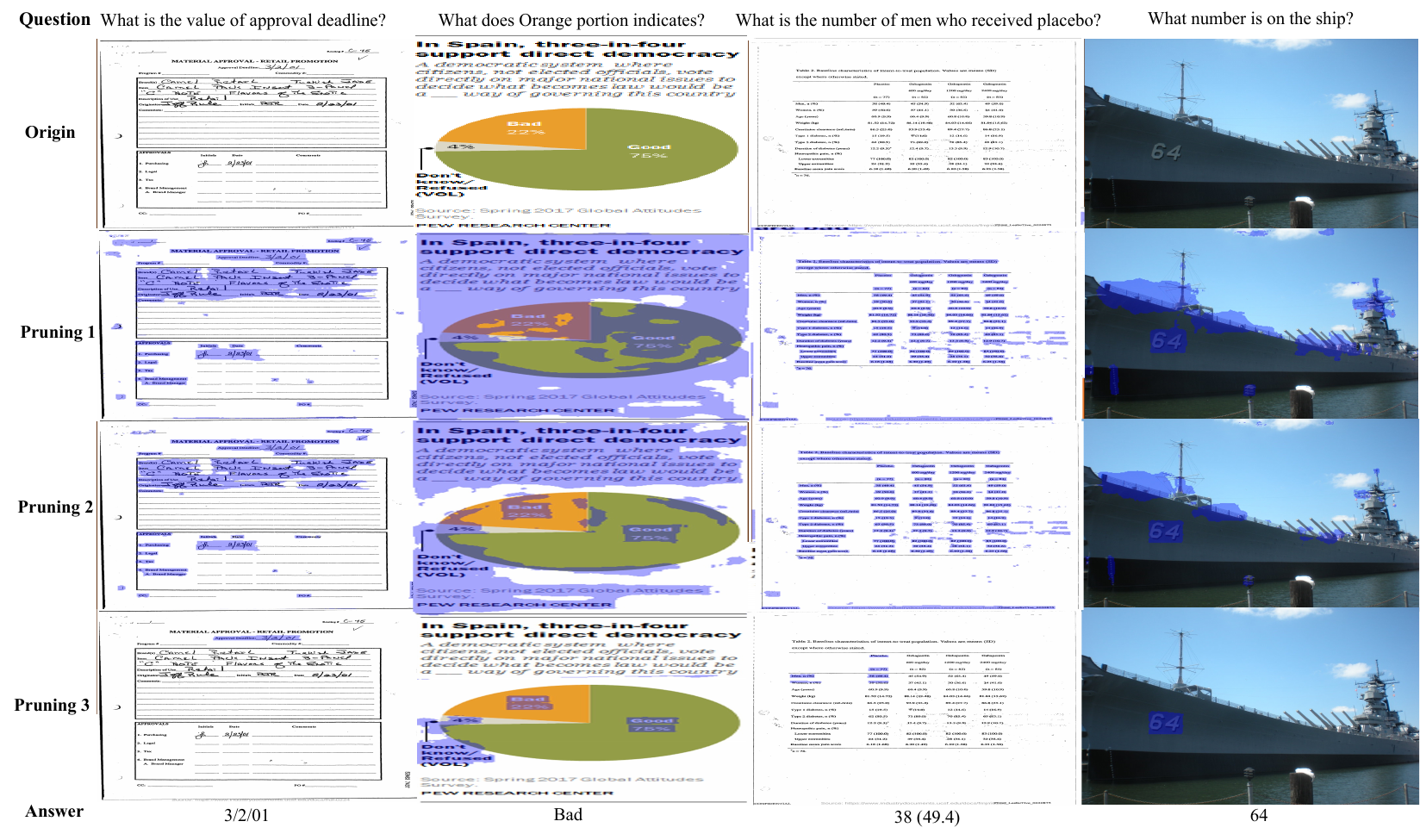}
   \caption{
    Visualization of the visual token filtering.
   The first row displays the original images, while the following three rows show the effects of visual token filtering. 
   The 
   % first two filtering stages 
   Pruning 1 and Pruning 2
   occur in the first two stages and the last two stages of the Swin Transformer, respectively, while Pruning 3 %the third filtering stage 
   takes place in the instruction filtering module.
   % Note that the filtered tokens are not completely discarded.
   % For aesthetic purposes, images are forcibly stretched to a consistent aspect ratio.
   }
   \label{fig:pruning}
   \vspace{-0.5cm}
\end{figure*}

\begin{figure}[ht]
  \centering
  \setlength{\abovecaptionskip}{0.02cm}
   \includegraphics[width=\linewidth]{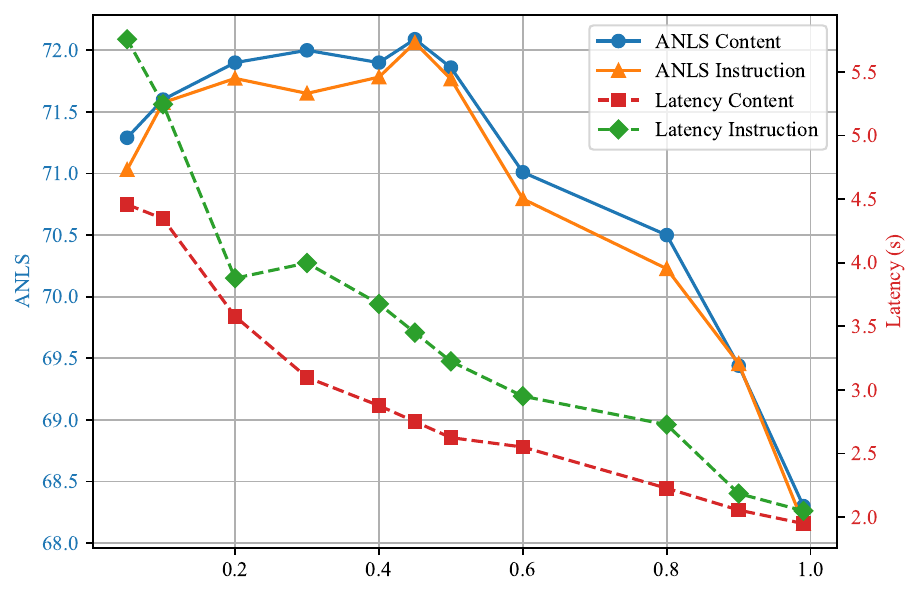}
   \caption{
   The impact of filtering thresholds on the DocVQA dataset.
   Best viewed in color.
   }
   \label{fig:f1_l}
   \vspace{-0.5cm}
\end{figure}

\begin{figure*}[ht]
  \centering
  \setlength{\abovecaptionskip}{0.02cm}
   \includegraphics[width=\linewidth]{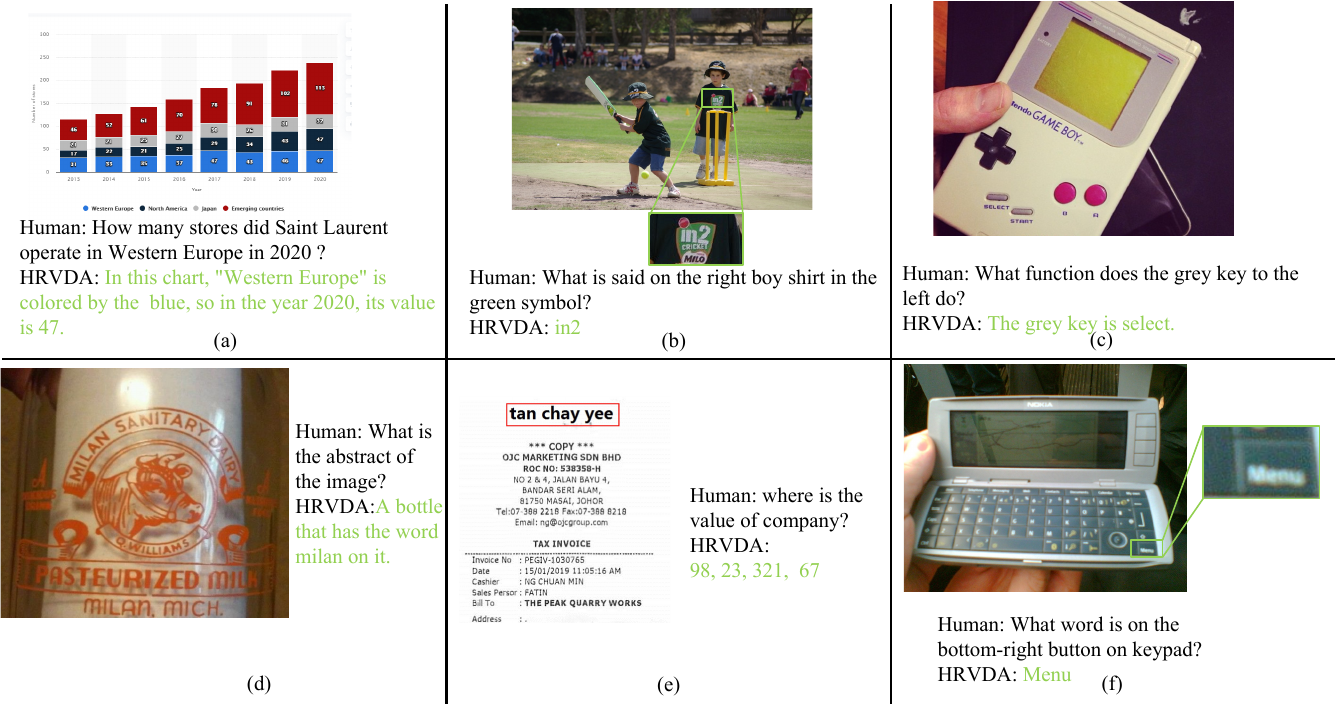}
   \caption{
   Qualitative examples generated by HRVDA. 
   For better clarity, key regions are magnified and cropped.
   }
   \label{fig:cases}
   \vspace{-0.3cm}
\end{figure*}

\section{Experiments}
\label{sec:exp}

In this section, we conduct experiments on numerous publicly available document-oriented datasets to validate the effectiveness of our proposed HRVDA model.

\subsection{Tasks and Datasets}
\label{sec:4.1}

In visual document understanding, information extraction and text-oriented visual question answering are challenging tasks, which also have widespread applications in practice.

\noindent
\textbf{Information Extraction} involves extracting structured key-value pair data from documents.
In this study, we use the two most commonly used datasets for evaluation, CORD~\cite{cord} and SROIE~\cite{sroie}. 
% Both are small-scale datasets, with the number of test set images \blue{ranging from a few hundred}.
They are all scanned receipt images and have good image quality.
The F1 score is reported, which is the weighted harmonic mean of Precision and Recall.

\noindent
\textbf{Text-oriented Visual Question Answering} is a highly generalizable task, capable of addressing various problems through appropriate prompts. 
We evaluate HRVDA on a wide range of publicly available datasets, including DocVQA~\cite{docvqa}, InfoVQA~\cite{InfographicVQA}, TextVQA~\cite{textvqa}, ChartQA~\cite{ChartQA}, DeepForm~\cite{due-deepform}, KLC~\cite{KLC}, WTQ~\cite{WikiTableQuestions}, TableFact~\cite{TabFact}, VisualMRC~\cite{VisualMRC}, and TextCaps~\cite{textcaps}. 
Different metrics, including ANLS, CIDEr, Accuracy, and F1 Score are reported in accordance with the methodologies employed in previous works.
A detailed description can be found in the Appendix~\ref{A.dataset}.

\subsection{Implementation Details}
\textbf{Model Architecture.}
Our HRVDA model employs Swin-L~\cite{swinT} as the image encoder. Its layer and window sizes are set to 2, 2, 18, 2, and 10, respectively, with a patch size of $4\times 4$. Additionally, the image resolution is set to $1536\times 1536$. 
In this study, we conduct experiments based on LLaMA-2-7B~\cite{llama2}, which has a context length of 4096.

\noindent
\textbf{Training Details.}
We employ the Adam optimizer for each stage of training, with an initial learning rate of 1e-4. The learning rate schedule uses a linear warmup during the first 20\% of steps. 
For LoRA, we set the rank to 8. 
Unless otherwise specified, the detection thresholds for content filtering in the Swin Transformer are set to $\epsilon_c = [0.25, 0.25, 0.5, 0.5]$ in 4 stages, while the threshold for instruction filtering is set to $\epsilon_i = 0.5$.
The batch size is set at 128. 
All training is conducted on 128 Tesla V100 GPUs for 10 epochs.
% 训练 epoch
% 用过滤 分辨率可以多大，不用过滤要多大。
% 推理时间比较

\subsection{Comparisons with Previous Approaches}
We conduct a comparative analysis of HRVDA against OCR-free models, including Donut~\cite{donut}, SeRum~\cite{serum}, Pix2Struct~\cite{pix2struct}, Qwen-VL~\cite{qwen-vl}, mPLUG-Doc~\cite{mPLUG-DocOwl}, and UReader~\cite{ureader}, utilizing 12 publicly available datasets for evaluation.

These models can be broadly categorized into two classes: encoder-decoder models and MLLMs. 
The first class utilizes a cross-attention mechanism~\cite{transformer} to fuse image and text, resulting in computational efficiency for high-resolution image inputs while simultaneously requiring task-specific fine-tuning. 
The second class leverages LLMs, offering exceptional %language
understanding capabilities, but often unable to directly %accept
process high-resolution inputs.

As demonstrated in Table ~\ref{tab:main}, HRVDA achieves the best results across the 9 datasets.
In information extraction tasks, our model significantly surpasses current state-of-the-art performance, owing to our robust visual pretraining (Stage 2). 
In visual question answering tasks, understanding the question becomes crucial, particularly in datasets with a high prevalence of elements from natural scene~\cite{textvqa}.
The semantic analysis capabilities of the decoder in the first category are limited, which prevents them from achieving optimal performance.
Previous MLLMs are constrained by the visual information distortion caused by low-resolution image input, which also prevents them from achieving desirable results.
Consequently, our HRVDA model directly processes high-resolution image inputs, minimizing the loss of visual information and thereby delivering substantial performance enhancements.

In terms of efficiency evaluation, we use Qwen-VL as our baseline and evaluate the forward-inference latency on a Tesla V100 GPU. The results reveal that HRVDA's speed is significantly faster than Qwen-VL's across various filtering thresholds, as illustrated in Table~\ref{tbl:time}. Remarkably, when both thresholds are set to 0.5, HRVDA reduces the runtime by 61\%. 
However, due to the constraints of GPU memory usage, we do not further increase the resolution.

\subsection{Ablation Study}
In this section, we separately explore the impact of filtering thresholds in the visual filtering mechanism and instruction filtering module.

Figure~\ref{fig:pruning} showcases several examples of token pruning. 
It can be observed that for text-dense images, the proportion of filtered pixels is considerably high. 
In contrast, for images containing charts and natural elements, the filtering ratio is lower, as more visual semantic information is required for these types of images. On the other hand, we quantitatively evaluate the impact of  filtering thresholds in the content filtering mechanism and instruction filtering module on prediction accuracy and inference latency, as shown in Figure~\ref{fig:f1_l}.
As the threshold increases, the accuracy of the prediction gradually improves, reaching its peak at 50\% and then experiencing a decline.
% The inference latency almost linearly increases with the token keep ratio. 
The inference latency decreases almost linearly with the filtering threshold.
These results indicate that appropriate token pruning not only accelerates computation but also improves performance, as removing redundant information can reduce the difficulty for the model to extract key information.

\subsection{Qualitative Analyzes}

As shown in Figure ~\ref{fig:cases}, HRVDA can recognize text in specific areas based on location hints. 
This is extremely useful in practical applications, as people often describe vague locations to obtain information.
HRVDA also successfully identifies the highly blurred text \textit{Menu}, which may be due to the influence of visual semantic cues.
Utilizing comprehensive document-oriented visual instruction tuning, HRVDA exhibits outstanding capabilities in following document instructions.
More cases can be found in the Appendix~\ref{A.qual}.

\section{Conclusion}
In this work, we propose a new OCR-free multimodal large language model, HRVDA, which can directly accept high-resolution image inputs and is suitable for fine-grained prediction tasks.
To the best of our knowledge, HRVDA is the first MLLM to utilize the Swin Transformer as an encoder.
Additionally, we employ a content filtering mechanism and an instruction filtering module to alleviate the computational challenges brought about by high-resolution inputs.
% We construct a document-oriented visual instruction tuning dataset and implement a multi-stage training strategy to fully exploit our model's document image modeling capabilities.
Experimental results demonstrate that our HRVDA model achieves state-of-the-art results on a series of public datasets, while also exhibiting significantly faster speeds compared to previous MLLMs.
In the future, we will continue to investigate high-resolution challenges.% in MLLMs.
\section*{Acknowlegement}
% This research was supported by the National Key Research and Development Plan of China (Grant No.2022YFB3103100), the National Natural Science Foundation of China (Grant No. 62276245), and Anhui Provincial Natural Science Foundation (Grant No. 2008085J31).
This research was supported by the National Key Research and Development Program of China (Grant No. 2022YFB3103100), the National Natural Science Foundation of China (Grant No. 62276245).
\clearpage

{
    \small
    \bibliographystyle{ieeenat_fullname}
    \bibliography{main}
}

\clearpage
\clearpage
\setcounter{page}{1}
% \maketitlesupplementary
\section*{Appendix\centering}
\appendix

\section{Analysis of Query-based Feature Extraction}
\label{A.1}
\begin{figure}[ht]
  \centering
   \includegraphics[width=\linewidth]{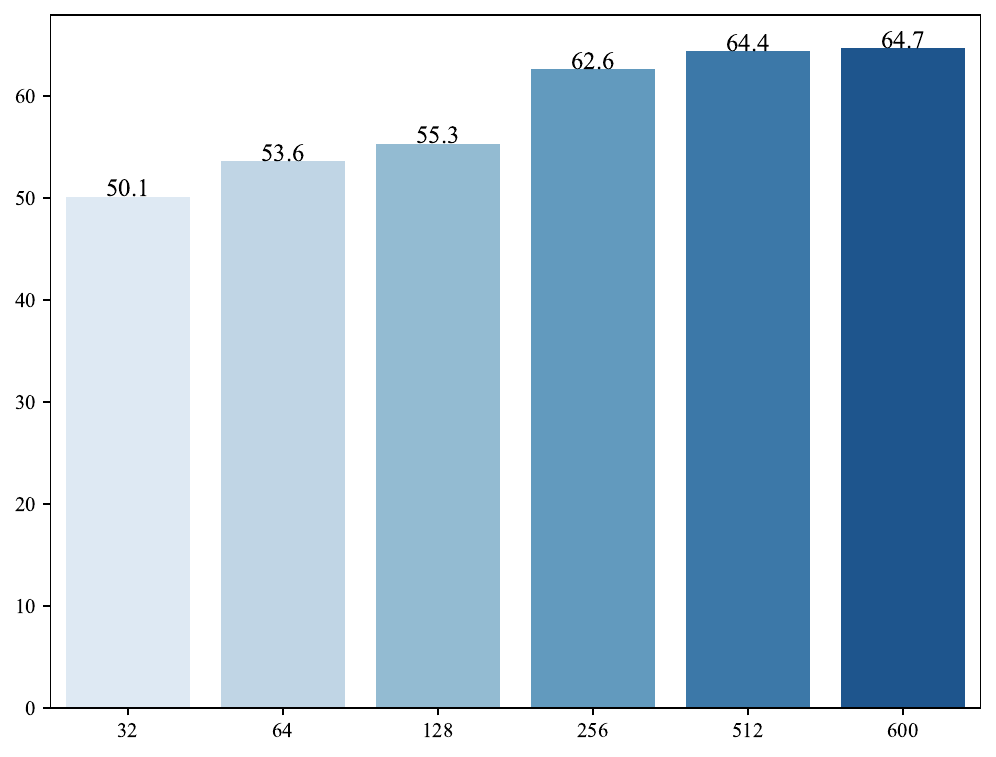}
   \caption{
   Performance with varying query numbers on the DocVQA dataset.
   }
   \label{fig:query}
   % \vspace{-0.5cm}
\end{figure}

In this section, we offer an experimental analysis to clarify our reasoning behind not choosing a query-based feature fusion approach.

Building upon the Donut framework~\cite{donut}, we employ Q-Former~\cite{blip} to extract image features and conduct cross-attention operations with Bart~\cite{mbart} using the extracted features. 
We fine-tune the model on the DocVQA dataset, and the experimental results are illustrated in Figure ~\ref{fig:query}. When the image resolution is set to 1280, we observe that an insufficient number of query vectors can significantly degrade the model's performance. To mitigate this decline while maintaining the model's performance, 500 query vectors are required. However, this approach to information extraction is not highly efficient in practice. 
Consequently, we choose a direct fusion approach in the instruction filtering module to retain visual information to the greatest extent possible.

\section{Visual Instruction Tuning}
\subsection{Instruction Templates}
\label{A.templates}
\begin{table}[t]
\centering
\begin{tabular}{ll}
\toprule
\textbf{Task}                 & \textbf{Format}                                     \\
\midrule
% \multirow{2}{*}{DC}  & Human: What is the category of this image? \\
%                      & AI: \{cls\}                                \\
% \hline
\multirow{8}{*}{IE}  & Human: What is the value of the \{key\}?   \\
                     & AI: \{value\}                              \\ 
                    & Human: What is the \{key\}?\\
                    & AI: \{value\} \\
                    & Human: What is the content of \{key\}?\\
                    & AI: \{value\} \\
                    & Human: What is the essence of the \{key\}?\\
                    & AI: \{value\} \\
                     \hline
% \multirow{2}{*}{VQA} & Human: \{question\}                        \\
%                      & AI: \{answer\}                             \\ \hline
\multirow{8}{*}{OCR} & Human: Present all the text in the image.  \\
                     & AI: \{all text\}                       \\        
                     & Human: please output the OCR result \\
                     & AI: \{all text\} \\
                     & Human: What is the text content in this image? \\
                     & AI: \{all text\} \\
                     & Human: What is the textual context of this image? \\
                     & AI: \{all text\}\\
                     \hline
\multirow{6}{*}{VG}  & Human: Where is the \{obj\}?               \\
                     & AI: \{x, y, x + w, y + h\}                 \\ 
                     & Human: Where is the \{obj\} recorded? \\
                     & AI: \{x, y, x + w, y + h\}  \\
                     & Human: Where is the \{obj\} located? \\
                     & AI: \{x, y, x + w, y + h\} \\
                     
                     \hline
\multirow{6}{*}{IC}  & Human: What is the abstract of the image?  \\
                     & AI: \{caption\}                            \\ 
                     & Human: Can you describe the content of this picture? \\
                     & AI: \{caption\} \\
                     & Human: Could you put into words what's in this picture ? \\
                     & AI: \{caption\} \\
                     & Human: Can you summarize this picture in one sentence?\\
                     & AI: \{caption\}\\
                     \hline

\multirow{4}{*}{TR}  & Human: What is the element in the table?   \\
                     & AI: \{element\}                           \\
                     & Human: Please output the table in kv format?\\
                     & AI : \{element\} \\
                     
\bottomrule
\end{tabular}
\setlength{\abovecaptionskip}{0cm}
% \setlength{\belowcaptionskip}{-0.2cm}
%\caption{Example demonstrations of instruction tuning templates tailored to seven tasks.}
\caption{%Example demonstrations 
% \red{Illustrative examples} of instruction tuning templates %tailored to
% \red{customized for} seven tasks.
Additional examples of instruction tuning templates. %tailored to
}
\label{tab:ins}
% \vspace{-0.5cm}
\end{table}

As shown in Table~\ref{tab:ins}, we present additional instruction templates. A greater number of instruction templates can significantly enhance the model's generalization capabilities and improve its performance in real-world applications. It is worth noting that users' perspectives in posing questions are diverse; therefore, having an adequate number of templates allows the model to better understand and respond to real-world instructions.

\subsection{Details of Datasets}
\label{A.dataset}
In this section, we provide a detailed introduction to the various datasets used in our experiments.

\noindent
\textbf{CORD} \quad
 The CORD~\cite{cord} dataset comprises 800 training receipts, 100 validation receipts, and 100 test receipts. Each receipt is accompanied by a photo and a set of OCR annotations. The dataset identifies 30 fields across four categories, and the task's objective is to correctly assign each word to the appropriate field. The evaluation metric used is the entity-level F1 score, and official OCR annotations are utilized.

\noindent
\textbf{SROIE} \quad
The SROIE~\cite{sroie} dataset is designed for extracting data from digitized receipts. It consists of 626 training samples and 347 testing samples. The objective is to retrieve information for up to four specific keys from each receipt: company, date, address, and total. The assessment metric used is the entity-level F1 score. Official OCR annotations are utilized, and the test set outcomes are supplied by the authorized evaluation platform.

\noindent
\textbf{DocVQA} \quad
The DocVQA~\cite{docvqa} dataset comprises 50,000 questions based on more than 12,000 pages from a diverse range of documents. The pages are divided into training, validation, and test sets at a ratio of approximately 8:1:1. The task's evaluation employs an edit distance-based metric called ANLS (average normalized Levenshtein similarity). 

\noindent
\textbf{InfoVQA} \quad
The InfographicVQA~\cite{InfographicVQA} dataset consists of 30,035 questions and 5,485 images, originating from 2,594 distinct web domains. This dataset employs the ANLS  metric for evaluation, where higher scores are assigned if the predicted answer has a smaller difference from at least one of the ground-truth answers.

\noindent
\textbf{DeepForm} \quad
DeepForm~\cite{due-deepform} is a socially important documents related to election spending with the objective of extracting contract numbers, advertiser names, payment amounts, and advertisement broadcast dates from advertisement disclosure forms. The dataset comprises 700 training samples, 100 validation samples, and 300 testing samples. The evaluation metric used is the F1 score.

\noindent
\textbf{KCL} \quad
Kleister Charity ~\cite{KLC} is a document understanding dataset designed for the extraction of information related to charitable organizations. It consists of 1,700 training samples, 400 validation samples, and 600 testing samples. The evaluation metric employed is the F1 Score.

\noindent
\textbf{WTQ} \quad
WikiTableQuestions~\cite{WikiTableQuestions} is a question-answering dataset that comprises semi-structured HTML tables sourced from Wikipedia. It includes 1,400 training samples, 300 validation samples, and 400 testing samples. The evaluation metric employed is accuracy.

\noindent
\textbf{TabFact} \quad
TabFact~\cite{TabFact} is a dataset designed for investigating fact verification tasks in the context of semi-structured evidence. It consists of 13.2K training samples, 1.7K validation samples, and 1.7K testing samples. The evaluation metric employed is accuracy.

\noindent
\textbf{ChartQA} \quad
ChartQA~\cite{ChartQA} is a question-answering dataset targeting data visualizations in the form of charts, involving both visual and logical reasoning. It comprises 9.6K manually curated questions and 23.1K questions automatically generated from manually curated chart summaries. 
 The evaluation metric employed is relaxed accuracy.

\noindent
\textbf{TextVQA} \quad
The TextVQA~\cite{textvqa} dataset is constructed by extracting images and questions from the Open Images v3 dataset. It consists of 34,602 training samples, 5,000 validation samples, and 5,734 testing samples. The evaluation metric employed is accuracy.

\noindent
\textbf{VisualMRC} \quad
The VisualMRC~\cite{VisualMRC} dataset aims to enable machines to read and comprehend text in real-world documents and respond to natural language questions. This dataset comprises over 30,000 question and abstractive answer pairs derived from more than 10,000 document images spanning multiple web domains. The evaluation metric employed is CIDEr.
The computation of CIDEr is based on syntactic consistency, content consistency, consistency metrics, and diversity evaluation, synthesizing the similarity and consistency scores between the generated image descriptions and multiple reference descriptions.

\noindent
\textbf{TextCaps} \quad
The TextCaps~\cite{textcaps} dataset consists of 28,408 images and 142,040 captions, requiring models to read and comprehend textual information within the images and generate coherent descriptions. The evaluation metric employed is CIDEr.

\section{Training}
\label{A.train}
\begin{figure}[t]
  \centering
   \includegraphics[width=\linewidth]{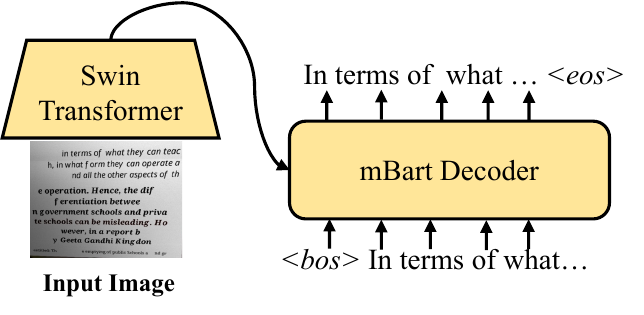}
   \caption{
   Schematic representation of the pretraining process for the image encoder.
   }
   \label{fig:donut}
   % \vspace{-0.5cm}
\end{figure}

In this section, we primarily provide a detailed description of Stage 2 of our training strategy.

Stage 2 essentially involves the pretraining of the image encoder. Currently, open-source image encoders mainly focus on two aspects: one is performing image classification tasks using datasets like ImageNet, and the other is aligning image and text features based on contrastive learning. These two pretraining paradigms are not suitable for generative tasks such as text recognition, as there is a significant difference between the pretraining methods and downstream tasks.

To make the image encoder more suitable for text recognition and generation tasks, we employ a method similar to Donut for pretraining the image encoder, as illustrated in Figure~\ref{fig:donut}. We primarily construct a temporary model to perform a pseudo-OCR task, which involves recognizing all text in the image in a top-to-bottom and left-to-right order. This pretraining task is more consistent with the downstream tasks, enabling our final HRVDA model to possess strong text recognition capabilities.

\begin{figure*}[t]
  \centering
   \includegraphics[width=\linewidth]{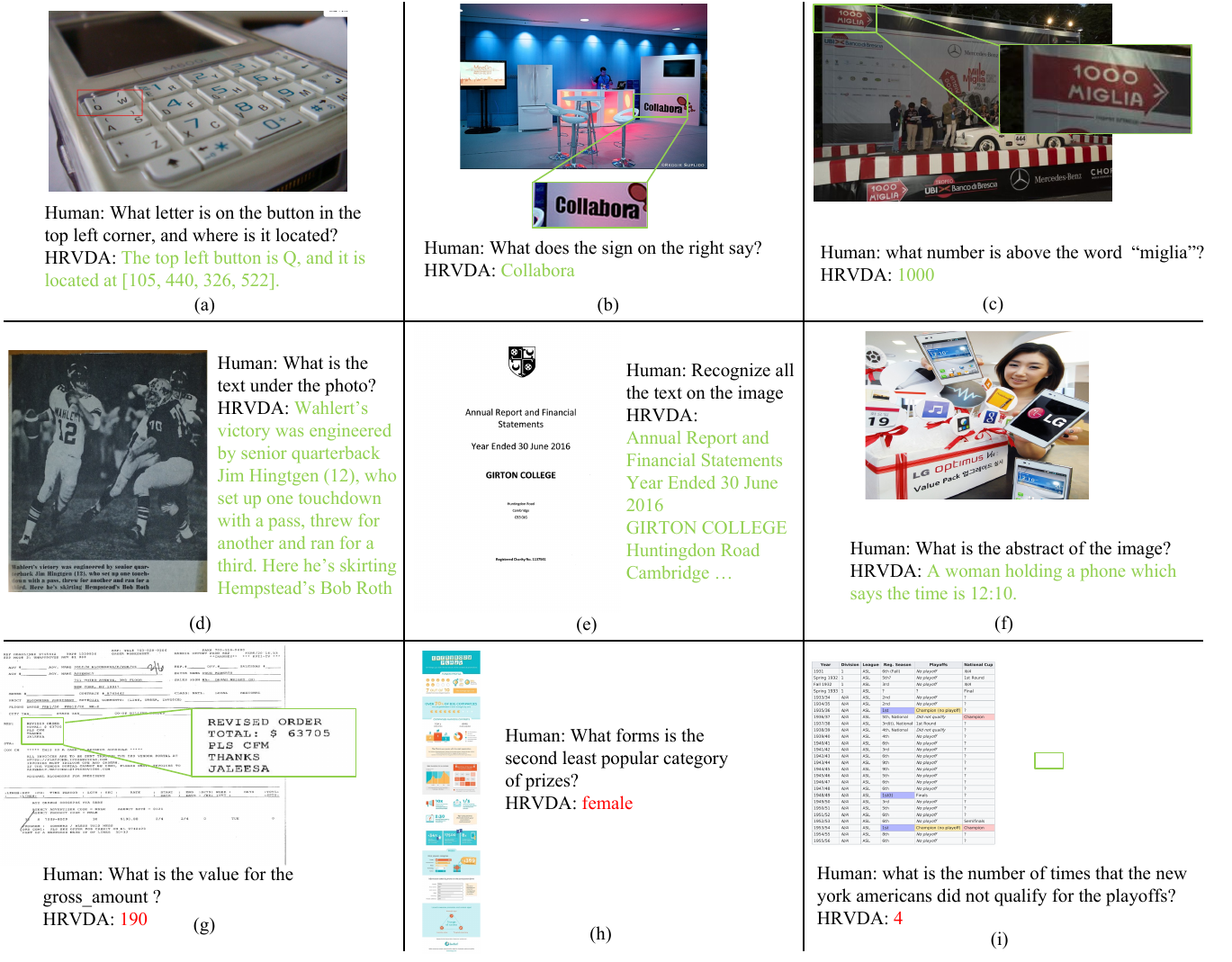}
   \caption{
   Additional qualitative examples generated by HRVDA.
   Green indicates that HRVDA answered correctly, while red represents incorrect answers.
   }
   \label{fig:case2}
   % \vspace{-0.5cm}
\end{figure*}

\begin{figure*}[t]
  \centering
   \includegraphics[width=\linewidth]{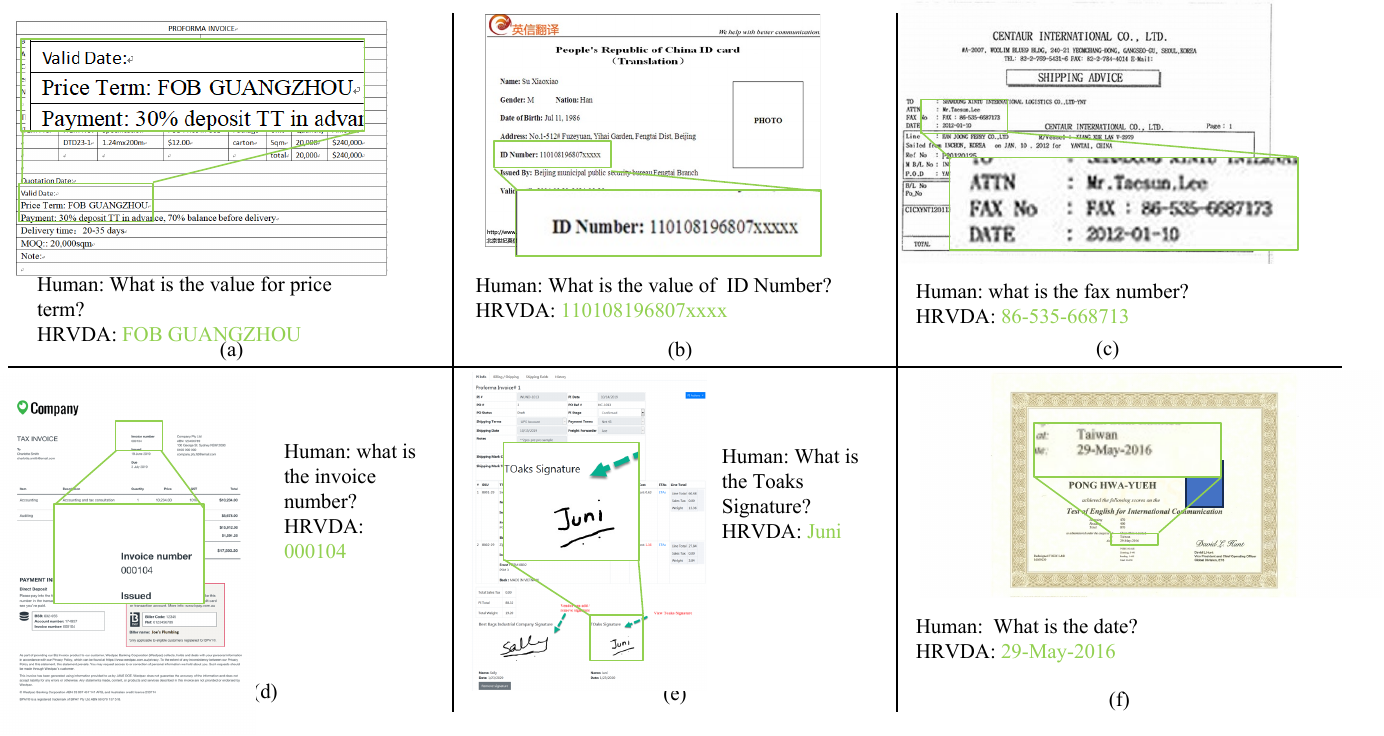}
   \caption{
 Performance demonstration of HDVDA on open-world examples.
   }
   \label{fig:opencase}
   % \vspace{-0.5cm}
\end{figure*}

\section{Qualitative Experimental Analysis}
\label{A.qual}
In this section, we provide some supplementary qualitative analysis.

% HRVDA possesses strong OCR capabilities. 
As depicted in the first two rows of Figure~\ref{fig:case2}, HRVDA can recognize colors, positions, and artistic fonts, which is primarily attributed to its  visual pretraining.
Furthermore, leveraging the semantic understanding capabilities of the LLM, HRVDA can also recognize text in complex regions, such as identifying the field above a particular field. 
Even when dealing with images containing long text, HRVDA demonstrates strong full-text OCR capabilities.

% However, perhaps due to the limited semantic capabilities of LLaMA-2-7B, HRVDA struggles with more complex instructions, such as those in Figure~\ref{fig:case2}-(f).
% However, HRVDA is unable to handle some particularly challenging examples, as shown in the second row of Figure~\ref{fig:case2}.
% For instance, HRVDA encounters comprehension errors when dealing with images that are particularly text-dense and possess complex structural relationships. 
% On the other hand, images with extreme proportions are not very friendly to HRVDA either. As shown in Figure~\ref{fig:case2}-(e), HRVDA can only handle such images with extreme proportions by attempting multiple cropping operations, but this directly causes the model to lose its understanding of the entire image structure. 
% Additionally, HRVDA is unable to generate a corresponding understanding for extremely complex instructions.
% To enhance the model's ability to handle challenging examples, we plan to increase the image resolution and employ a more powerful LLM in the future.

Nonetheless, HRVDA struggles with certain highly challenging examples, as illustrated in the last row of Figure~\ref{fig:case2}. 
For example, the HRVDA model faces comprehension difficulties when processing images that have an exceptionally high density of text and exhibit intricate structural relationships. Moreover, HRVDA is not well-suited for images with extreme proportions. As demonstrated in Figure~\ref{fig:case2}-(h), the model can only manage such images by performing multiple cropping operations, which inevitably compromises its grasp of the overall image structure. Furthermore, HRVDA is incapable of generating an adequate understanding for exceedingly complex instructions. 
To address these extremely challenging examples, we plan to further increase the resolution and employ a more powerful LLM in future iterations.

% Considering the absolute performance across various datasets, the model's performance is somewhat lacking for datasets that have a high dependency on instruction semantic understanding.

% We also evaluate the performance of HRVDA using open-domain data, as shown in Figure~\ref{fig:opencase}.
% HRVDA excels at handling various information extraction tasks, significantly advancing the practical implementation of MLLMs. 

We also evaluate the performance of HRVDA using open-domain data, as shown in Figure~\ref{fig:opencase}.
HRVDA performs exceptionally well in information extraction tasks for common fields, such as dates, amounts, fax numbers, \textit{etc}.
Overall, if the answer relies more on simple text recognition, HRVDA can perform very well, significantly advancing the practical application of MLLMs.

% \clearpage
% \begin{figure*}[t]
%   \centering
%   % \fbox{\rule{0pt}{2in} \rule{0.9\linewidth}{0pt}}
%    \includegraphics[width=\linewidth]{case2.pdf}
%    \caption{
%    Additional qualitative examples generated by HRVDA.
%    Green indicates that HRVDA answered correctly, while red represents incorrect answers.
%    }
%    \label{fig:case2}
%    \vspace{-0.5cm}
% \end{figure*}

% \begin{figure*}[t]
%   \centering
%   % \fbox{\rule{0pt}{2in} \rule{0.9\linewidth}{0pt}}
%    \includegraphics[width=\linewidth]{opencase.pdf}
%    \caption{
%  Performance demonstration of HDVDA on open-world examples.
%    }
%    \label{fig:opencase}
%    \vspace{-0.5cm}
% \end{figure*}


\end{document}